\pdfoutput=1

\documentclass[11pt]{article}

\usepackage[preprint]{acl}

\usepackage{times}
\usepackage{latexsym}

\usepackage[T1]{fontenc}
\usepackage[utf8]{inputenc}

\usepackage{microtype}

\usepackage{inconsolata}

\usepackage{comfortaa}

\usepackage{ulem}

\usepackage{listings}
\usepackage{xcolor}
\usepackage{tcolorbox}
\usepackage{graphicx}

\usepackage{array}
\usepackage{booktabs}

\DeclareRobustCommand{\augiefamily}{%
  \fontfamily{augie}\fontseries{m}\fontshape{n}\selectfont}
\DeclareTextFontCommand{\textaugie}{\augiefamily}

\newcommand{\sxm}[1]{}

\title{Whiteboard-of-Thought: Thinking Step-by-Step Across Modalities}

\author{Sachit Menon \\
  Columbia University\\
  \texttt{sachit.menon@columbia.edu} \\\And
  Richard Zemel \\
  Columbia University \\
  \texttt{zemel@cs.columbia.edu} \\
  \\\href{http://whiteboard.cs.columbia.edu}{\texttt{whiteboard.cs.columbia.edu}}
  \And
  Carl Vondrick \\
  Columbia University \\
  \texttt{vondrick@cs.columbia.edu} \\
  }

\begin{document}
\maketitle
\begin{abstract}
When presented with questions involving visual thinking, humans naturally switch reasoning modalities, often forming mental images or drawing visual aids. Large language models have shown promising results in arithmetic and symbolic reasoning by expressing intermediate reasoning in text as a chain of thought, yet struggle to extend this capability to answer text queries that are easily solved by visual reasoning, even with extensive multimodal pretraining. We introduce a simple method, \textit{whiteboard-of-thought} prompting, to unlock the visual reasoning capabilities of multimodal large language models across modalities. Whiteboard-of-thought prompting provides multimodal large language models with a metaphorical `whiteboard' to draw out reasoning steps as images, then returns these images back to the model for further processing. We find this can be accomplished with no demonstrations or specialized modules, instead leveraging models' existing ability to write code with libraries such as Matplotlib and Turtle. This simple approach shows state-of-the-art results on four difficult natural language tasks that involve visual and spatial reasoning. We identify multiple settings where GPT-4o using chain-of-thought fails dramatically, including more than one where it achieves $0\%$ accuracy, while whiteboard-of-thought enables up to $92\%$ accuracy in these same settings. We present a detailed exploration of where the technique succeeds as well as its sources of error.
\end{abstract}

\section{Introduction}

Which lowercase letter is a circle with a vertical line touching it to the right going down?

This \textbf{\comfortaa{q}}uestion may sound rather trivial. %
Likely, you solved it by forming a mental image as you proceeded through the sentence, first placing the circle then adding the line to recognize the letter `q.' If there had been more pieces to keep track of, you might instead resort to pen and paper, but follow a similar process. 

\begin{figure*}[ht!]
    \centering
    \includegraphics[width=\linewidth]{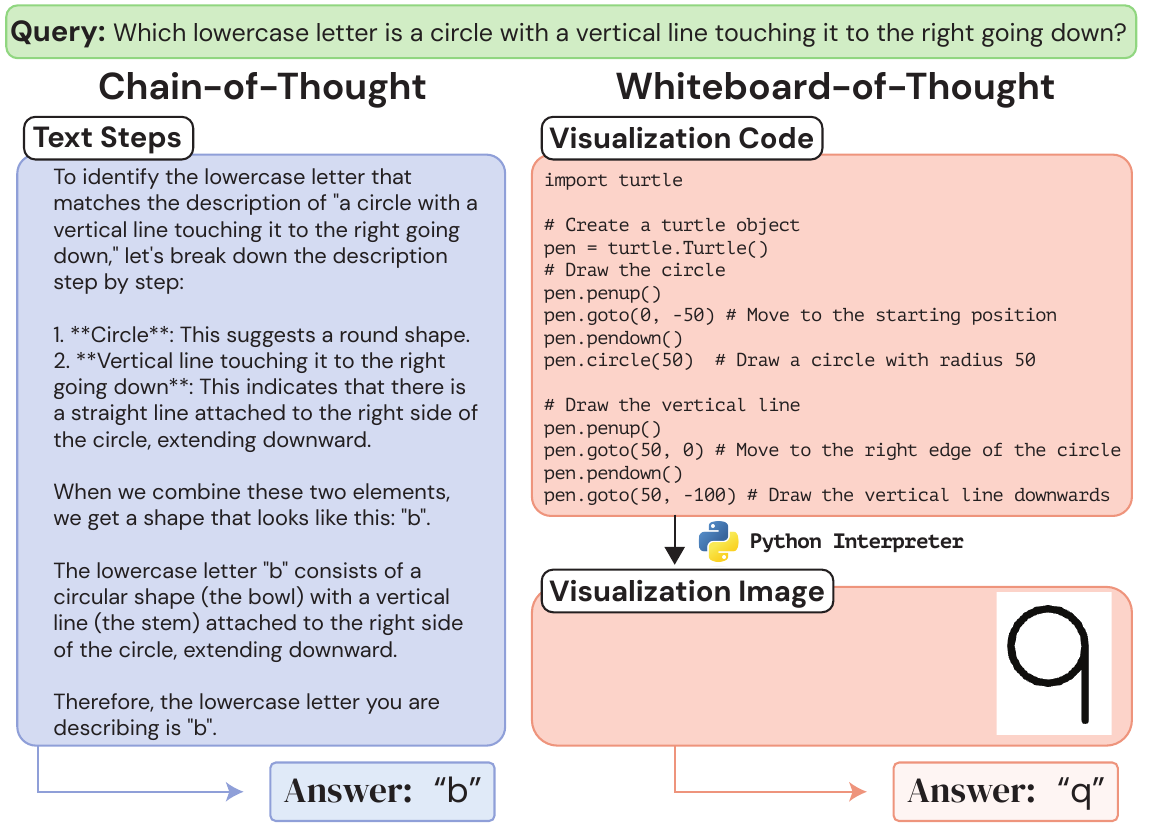}
    \caption{For queries that are trivial with visual reasoning, chain-of-thought (which produces the text on the left) can fail in surprising ways. 
    Whiteboard-of-thought (which produces the code, image, and text on the right) provides an alternative to perform intermediate reasoning with images.}\vspace{-1em}
    \label{fig:cot}
\end{figure*}

\looseness-1 Humans excel at this type of \textit{visual thinking}. We readily interweave reasoning in words and reasoning in images to solve problems and communicate ideas \cite{tversky_visualizing_2011}; we form images not only for directly visual reasoning, but also to draw maps during spatial navigation \cite{card_readings_1999} and even process abstract concepts \cite{bobek_creating_2016}. 

Large language models (LLMs) have revolutionized NLP \cite{brown_language_2020,vaswani_attention_2017}, showing the power of scaling parameter count and training data in all sorts of reasoning tasks. 
Chain-of-thought (CoT) prompting \cite{wei_chain_2022,kojima_large_2023}, which allows a language model to solve complex problems by writing out intermediate steps as text before producing a final answer, and similar techniques have shown huge promise in tasks such as arithmetic and symbolic reasoning. 
It is thus natural to ask: can LLMs solve tasks that we, as humans, solve with visual thinking?

We find that for some tasks involving visual and spatial reasoning, even the best LLMs fail dramatically. In Fig.~\ref{fig:cot}, we see GPT-4o \cite{openai_gpt-4_2023} claim 
the answer to our earlier question is in fact `b'; it does not perform the correct implicit visual reasoning about left, right, up, and down to reach the correct answer.
`b' was observed sampling deterministically; with temperature, it picks randomly among `b', `p', `d', and `q', i.e., the letters with the correct features but in different arrangements.
It references the `bowl' and the `stem' to narrow down potential options then fails to perform the spatial reasoning to determine what the described orientation \textit{means}. A formal background in typography should not be needed to answer the question; anyone who can create the correct visual with a basic level of visual recognition will solve it easily. This highlights the significant differences between the two reasoning processes. %

Our key idea is that visual reasoning tasks demand visuals. %
We leverage the abilities of multimodal large language models (MLLMs), which accept inputs not only in text but other modalities such as images, to achieve this. We show that providing MLLMs the ability to create and reason with explicit visuals -- like a whiteboard showing intermediate thoughts -- unlocks capabilities resembling visual thinking. %

We thus introduce `\textbf{whiteboard-of-thought} (WoT)': we provide MLLMs with a metaphorical `whiteboard' to draw out the results of intermediate reasoning steps as \textit{images}, then prompt them to use their visual input capabilities to produce answers or perform further reasoning from the visuals made by the model itself. We find that leveraging models' existing ability to write code with visual graphics libraries such as Turtle and Matplotlib proves sufficient to create visuals useful for solving visual reasoning tasks without requiring a single example.

We demonstrate the potential of this idea on three BIG-Bench \cite{srivastava_beyond_2022} tasks involving understanding ASCII art, as well as a recent difficult benchmark probing spatial reasoning abilities \cite{yamada_evaluating_2024}, establishing a large perfomance gap between WoT and CoT. We further breakdown what types of problems are best suited to performing reasoning on visual tokens rather than text tokens. Finally, we identify current limitations of MLLM abilities and provide a detailed analysis of where WoT fails.

\section{Preliminaries}

We provide a brief overview of the main concepts underlying this work here. %

\paragraph{LLMs and MLLMs}
Large language models have seen substantial success across a wide range of natural language tasks by scaling data and parameter counts \cite{brown_language_2020,vaswani_attention_2017}. Recent work has extended advances in language modeling to the multimodal input setting, leading to multimodal large language models (MLLMs). These models generate text conditioned on not other text as context, but also inputs from other modalities, enabling tasks such as captioning or visual question answering. Of particular note for this work, the large amount of image-and-text data available on the Internet has enabled this effort to prove especially successful for image inputs, achieving state-of-the-art results across a wide range of computer vision tasks \cite{li_blip-2_2023,alayrac_flamingo_2022,liu_visual_2023}. 

\paragraph{Chain of thought prompting}
Despite the successes of scaling for large language models, directly producing answers to complex, multi-step reasoning tasks, such as arithmetic or symbolic tasks, historically has remained difficult. \citet{wei_chain_2022} introduced a simple technique to substantially improve LLMs' performance on these tasks: few-shot chain-of-thought, prompting the model to break down complex queries and answer step-by-step by providing examples of the desired step-by-step reasoning as context. \citet{kojima_large_2023} broke this reliance on in-context exemplars to develop zero-shot chain-of-thought (in this work, referred to as simply `chain-of-thought' or CoT). Instead of using handwritten examples, they perform two simple steps to elicit step-by-step reasoning. First, they prompt the model with the query and an additional instruction to `think step by step'; the model then generates text comprising a reasoning trace. Second, they feed the reasoning trace back to the model, along with the instruction to produce an answer. This approach highlighted the substantial potential benefits of intermediate reasoning in the attractive zero-shot setting.

\paragraph{Tool augmented large language models}
Other work has shown that large language models can be induced to use external tools, such as calculators, to aid this intermediate reasoning. \citet{nye_show_2021} provide language models with a scratchpad: a dedicated buffer of text designated for intermediate computation, trained to mimic Python code execution. \citet{gao_pal_2023} and \citet{chen_program_2023} instead use LLMs' ability to write code to delegate simple computation to the Python interpreter, providing the computed results back to the LLM as text. To the best of our knowledge, we are the first to consider this ability's potential for generating visualizations to aid with intermediate reasoning.

\section{Whiteboard-of-Thought}

The goal of this work is to equip MLLMs with the ability to create images and visually process them to better answer queries. Our approach operates this whiteboard by synthesizing drawing code, executing this code to create the drawing, and parsing the resulting image before producing a final answer. Fig. \ref{fig:cot} shows an example of the full procedure.

\paragraph{Creating visuals with MLLMs.} Current MLLMs typically do not inherently possess the ability to produce outputs in the visual domain. Instead, we will show how we can create visuals using a model that only produces text.

The images we create for visual reasoning tend to be minimal, abstract, and symbolic \cite{tversky_visualizing_2011}. We use code as a natural way to create such visuals. Leveraging what models already know about common Python libraries like Matplotlib or Turtle enables this capability to emerge zero-shot, without needing any specialized, hand-crafted modules (though these could be made to adapt the technique to specific domains). We discuss other approaches such as text-to-image models in the Appendix.

In order to generate the code, we provide the MLLM with the query and prompt it to write code to visualize it. For each query, we prompt the MLLM with the input \texttt{You write code to create visualizations using the \{Matplotlib/Turtle\} library in Python, which the user will run and provide as images. Do NOT produce a final answer to the query until considering the visualization.} along with the query. The model then decides what visualization code to write based on the query. Full prompting and inference details for code generation can be found in the Appendix.

The resulting code is then passed to a runtime environment to render it in image form. In this case, we use the Python interpreter with the previously-mentioned visualization libraries.

\paragraph{Processing the generated visuals.} To process the resulting image, we make use of the MLLM's intrinsic multimodal input capacity. This obviates the need for external tools, like handcrafted visual modules \cite{gupta_visual_2022,suris_vipergpt_2023}, leading to a tightly self-contained method. %

In summary, the model is given a prompt including the query, the knowledge that it can use the aforementioned visualization libraries, and that it will be provided the resulting images along with the query to perform further steps towards producing a final answer. It produces code, which is then executed to create an image. The resulting visual is returned back to the model to perform the next step or produce an answer. %

\begin{figure}
    \centering
    \includegraphics[width=\linewidth]{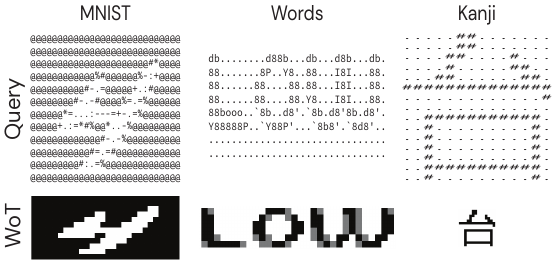}
    \caption{Example queries for each of the three ASCII understanding BIG-Bench tasks \cite{srivastava_beyond_2022} we consider, along with the WoT visualization for each.}
    \label{fig:ascii}
\end{figure}

\section{Experiments}

We conduct experiments in two broad classes of natural language tasks that involve visual reasoning. First, we consider three datasets from BIG-Bench \cite{srivastava_beyond_2022} that involve understanding information represented as ASCII text graphics.
Next, we consider tasks involving natural language navigation under different spatial conditions \cite{yamada_evaluating_2024}.

We perform all experiments in the zero-shot regime, and compare to two baselines without visualization: directly prompting the model for an answer (`Direct') and zero-shot chain-of-thought \cite{kojima_large_2023} (`CoT'). We use a temperature of 0 and greedy decoding for generation. For all experiments, we use GPT-4o (gpt-4o-2024-05-13) as the underlying MLLM as it has each of the necessary capabilities enabling our model and the baselines -- zero-shot chain-of-thought as well as the ability to produce code outputs and to accept image inputs. %
Full prompts and other generation details can be found in the Appendix.

\subsection{ASCII Understanding}

We begin with a clearly visual task found in BIG-Bench: ASCII understanding. Recent work has shown that even the strongest language models struggle to recognize ASCII representations, and that this failure can even be exploited to perform highly effective jailbreak attacks resulting in unintended and unsafe behaviors that bypass state-of-the-art defense techniques \cite{jiang_artprompt_2024}. 

ASCII art highlights our ability to subconsciously switch between processing modalities: it requires reinterpreting characters that usually possess some natural language interpretation (e.g., `=' as an equals sign) in a visual context, focusing on their arrangement and spatial relationships (e.g., `======` as a horizontal line). For humans, written text is typically processed with the same input modality as images (our eyes), allowing us to engage in visual thinking without any intermediate processing. %

Consider the difficulty of understanding ASCII art being read aloud. 
This can be thought of as similar to how LLMs process ASCII: as text tokens, distinct from any visual tokens they may be able to process if they have multimodal capabilities. %
Thus, ASCII presents an interesting testing ground for evidence of visual thinking in MLLMs.

We consider three domains of ASCII understanding each comprising a task in BIG-Bench \cite{srivastava_beyond_2022}: ASCII MNIST digit recognition, ASCII word recognition, and ASCII kanji (Japanese logographic character) recognition. 
Examples of each of these can be found in Fig. \ref{fig:ascii} (along with the WoT visualization for each). %
Dataset and evaluation details can be found in the Appendix. 

Results can be found in Table \ref{tab:ascii}. We find that state of the art MLLMs are largely incapable of performing visual representation on these textual inputs.
Prompting for step-by-step reasoning in words provides little benefit.
However, providing a whiteboard to enable models to create and consider their own visualizations unlocks the visual thinking abilities latent in the MLLM, leading to substantial performance improvements. %

\begin{table}[t]
\centering
\begin{tabular}{>{\bfseries}lccc}
\toprule
& \textbf{MNIST} & \textbf{Word} & \textbf{Kanji} \\
\midrule
Direct & 19.6 & 24.8 & 1.1 \\
CoT & 21.6 & 27.2 & 1.1 \\
\textbf{WoT (ours)} & \textbf{66.0} & \textbf{66.4} & \textbf{73.8} \\
\bottomrule
\end{tabular}
\caption{\textbf{ASCII recognition accuracy.} MLLMs fail to perform the task with text alone. WoT unlocks visual processing to achieve substantial gains.}
\label{tab:ascii}
\end{table}

\paragraph{Text-based approaches vs visualization}
At first glance, the baseline performance on the MNIST and word recognition tasks may seem surprisingly high, potentially giving the impression that with sufficient scale, text alone could conceivably solve these tasks. Upon deeper inspection of where the baselines succeed compared to where WoT succeeds, however, this notion quickly falls apart. 

\begin{figure}
    \centering
    \includegraphics[width=\linewidth]{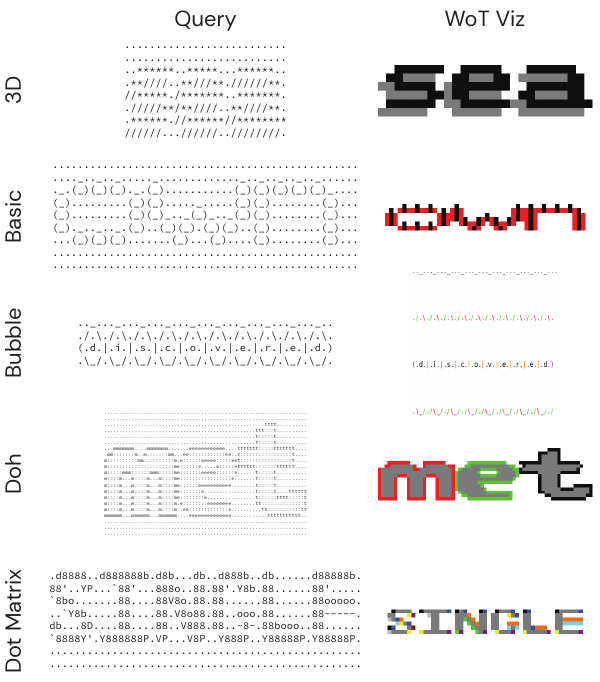}
    \caption{The different forms of ASCII in the BIG-Bench ASCII Word Recognition task and the visualizations made by WoT. Note that `Bubble' simply includes the word with some additional characters, and `Doh' forms the shape of each letter out of itself. (CoT results: `ascii', `hello', `discovered', `meet', `goodbye'.) Best viewed with zoom.}
    \label{fig:fonts}
\end{figure}

Examining the word recognition task, we find that the text-only baselines fail to solve \textit{every} instance that actually requires visual understanding, i.e., that does not directly use the characters from the word to form the ASCII string.
In more detail, the BIG-Bench ASCII word recognition task uses five forms of ASCII, shown in Fig. \ref{fig:fonts} along with visualizations made by WoT for each. Note that `Bubble' simply includes the word with some additional characters, and `Doh' forms the shape of each letter out of itself. In other words, they can be solved without considering the queries visually. 
We see in Table \ref{tab:font_comparison} that the only examples which the text-based approaches recognize come from these categories. 
The small boost CoT provides comes entirely from the `Doh' category, which can be solved with linguistic reasoning. On the other hand, WoT achieves dramatically higher performance on every category other than `Bubble', that being the only fully `non-visual' of the five.

\begin{table*}[ht]
\centering
\begin{tabular}{>{\bfseries}lccccc}
\toprule
& \textbf{3D} & \textbf{Basic} & \textbf{Bubble} & \textbf{Doh} & \textbf{Dot Matrix} \\
\midrule
Direct & 0.0 & 0.0 & \textbf{100.0} & 50.0 & 0.0 \\
CoT & 0.0 & 0.0 & \textbf{100.0} & 62.5 & 0.0 \\
\textbf{WoT (ours)} & \textbf{92.1} & \textbf{78.0} & 42.1 & \textbf{89.6} & \textbf{11.8} \\
\bottomrule
\end{tabular}
\caption{\textbf{ASCII word accuracy breakdown}. Using text alone fails \textit{every} instance of `visual' ASCII (see Fig. \ref{fig:fonts}).}
\label{tab:font_comparison}
\end{table*}

\paragraph{Does ASCII understanding necessitate `reasoning'?} One might contend that, for the ASCII understanding task, `constructing' a helpful visual could simply be rendering the ASCII directly as an image, i.e., a fixed procedure with no dynamic reasoning involved. At a motivational level, we are using these tasks to evaluate our more general-purpose method; the goal is not to devise an ASCII-specific technique. Nevertheless, we conduct an experiment comparing to this fixed baseline for the word recognition task. That is, rather than generating code on a per-query basis to create visualizations as in WoT, for this baseline we render the text directly as images and evaluate whether MLLMs can understand these images directly.%

We find that this approach results in an accuracy of $22.0\%$ -- comparable to (and in fact lower than) the text-only baseline. Why? We manually inspect the results, and find the errors fall into two broad qualitative categories. First, ASCII art rendered as text seems to result in some level of task confusion, similar to `typographic attacks' \cite{noever_reading_2021,menon_task_2022,goh_multimodal_2021,materzynska_disentangling_2022,ilharco_patching_2022}, between performing the intended task and the character-level OCR task. Second, many of the produced outputs are `technology'-related, for instance `hello world' or `Google'; we hypothesize this may stem from the style of ASCII writing may be spuriously correlated with these concepts on the Internet. At a high level, drawing yourself a confusing visual is worse than drawing no visual at all.

On the other hand, WoT allows the model to reason about how to render the visual. As seen in Fig. \ref{fig:fonts}, different types of ASCII input can be better served by different visuals. For instance, direct rendering may be suitable for `Bubble', while others such as `3D' can be made more legible -- even for humans -- with more careful visualization. %

\paragraph{Error analysis}
One of the benefits WoT presents is the ease of error attribution: that is, for each instance, was the error due to issues in producing the visualization, such as code execution errors or incorrect visualization, or issues in visual recognition from the generated visuals?

As the ASCII MNIST dataset is originally derived from actual images, it provides an avenue to `ground truth visualizations.' By measuring the performance of the MLLM on the real images as an `upper bound,' we can obtain insight into how much of the error can be attributed to each of these causes. If the model were able to produce the `ideal' visualization for every data point, how much would its performance be limited by its visual perception capabilities? We find that the MLLM obtains an accuracy of $80.8\%$ on actual MNIST images. This suggests a substantial proportion of the remaining error can be attributed to perception. 

\begin{figure}
    \centering
    \includegraphics[width=\linewidth]{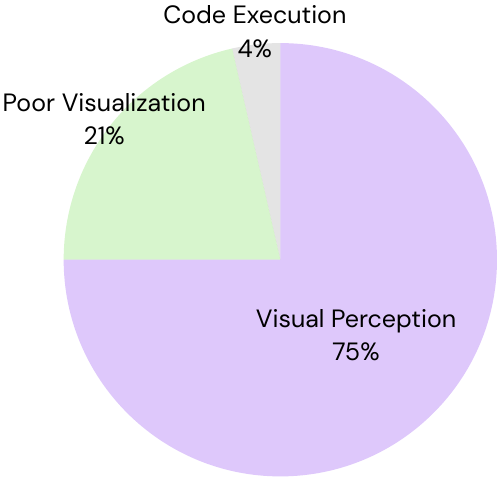}
    \caption{A qualitative breakdown of the sources of error for WoT evaluated on the ASCII MNIST task.}
    \label{fig:mnisterror}
\end{figure}

To understand the remaining sources of error for WoT further, we conducted an in-depth error analysis of the results on the ASCII MNIST task. For each error, we qualitatively categorized each error by its apparent cause, finding three broad categories. First, if no image was produced at all, e.g. due to the visualization script raising an error, we labeled the cause `code execution.' Otherwise, the authors determined if they would find it easy to produce the correct answer from the generated image, judging it a visual perception error if so and poor visualization if not. 
The results of this analysis can be found in Fig. \ref{fig:mnisterror}. We find that indeed, the errors largely stem from visual perception. 

\subsection{Spatial Navigation}

\begin{figure}
    \centering
    \includegraphics[width=\linewidth]{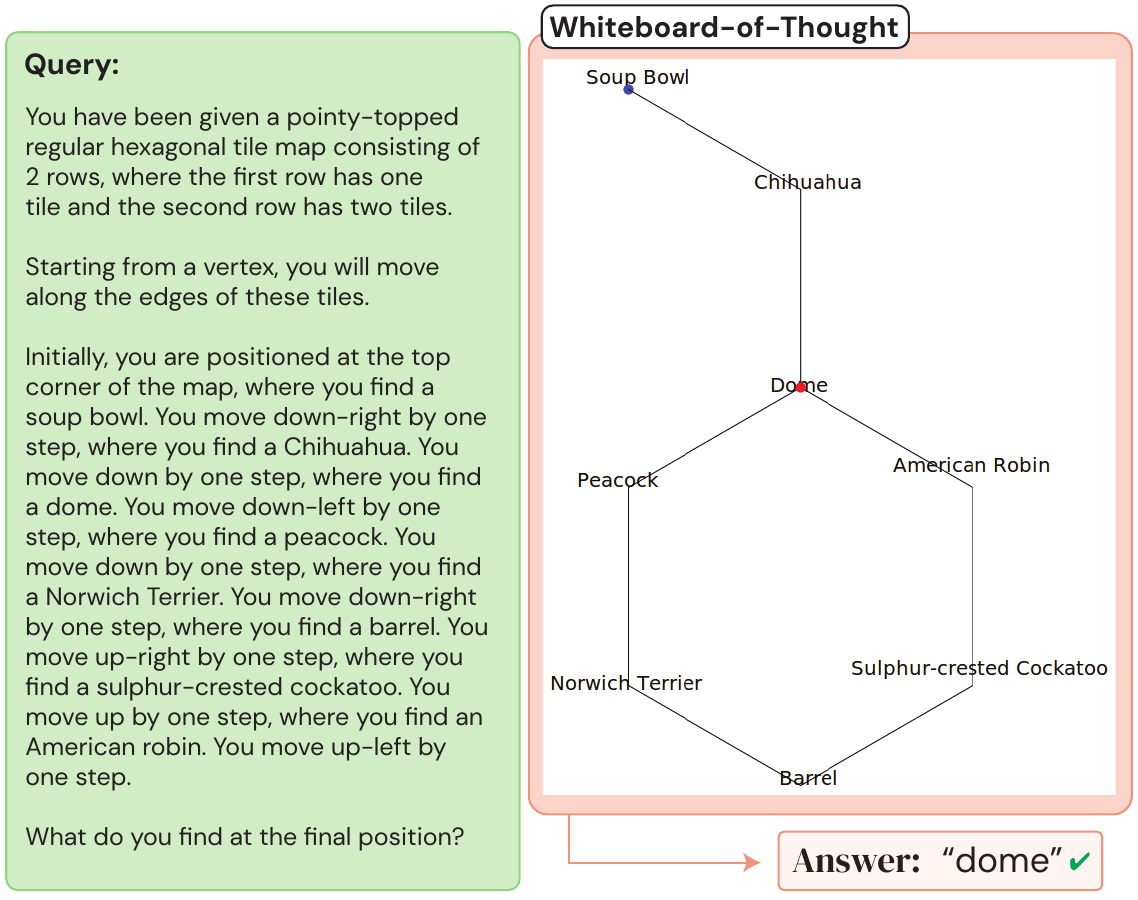}
    \caption{Example WoT visual for spatial navigation.}
    \label{fig:spatial}
\end{figure}

Next, we consider the task of understanding the spatial implications of natural language navigation instructions. Given a sequence of spatial instructions like in Fig. \ref{fig:spatial}, humans typically solve these tasks with visual thinking, such as creating a mental image or drawing a physical map \cite{garvert_map_2017,tversky_visualizing_2011,bobek_creating_2016}. We aim to understand whether MLLMs can solve these tasks in text alone, possibly suggesting some level of implicit visual thinking for spatial navigation, or if providing a whiteboard to draw an explicit map can provide additional value. 

A simple navigation task appeared in BIG-Bench \cite{srivastava_beyond_2022}, but only considers movement forwards and backwards in a straight line.%
The recent work of \citet{yamada_evaluating_2024} presents a more complex evaluation suite for probing the spatial understanding of LLMs, including navigation across a variety of spatial structures. In particular, we note a distinction between the 2D grid navigation setting -- `Square' and `Rhombus', here defined as the square rotated 45 degrees -- and the non-grid geometries (`Circle', `Hexagon', and `Triangle'). We therefore elect to consider this evaluation suite. We present an example, along with the WoT visualization result, in Fig. \ref{fig:spatial}. Details can be found in the Appendix. %

\begin{table*}[ht]
\centering
\begin{tabular}{lccccccc}
\toprule
& \multicolumn{3}{c}{\textbf{Non-Grid Structures}} & \multicolumn{2}{c}{\textbf{2D Grids}} & \textbf{Overall} \\
\cmidrule(lr){2-4} \cmidrule(lr){5-6} \cmidrule(lr){7-7}
& \textbf{Circle} & \textbf{Hexagon} & \textbf{Triangle} & \textbf{Square} & \textbf{Rhombus} & \textbf{Avg} \\
\midrule
Direct & 14 & 3 & 16 & 68 & \textbf{63} & 33 \\ %
CoT & 25 & 8 & 26 & \textbf{98} & 51 & 42 \\ %
\textbf{WoT (ours)} & \textbf{41} & \textbf{61} & \textbf{55} & 50 & 52 & \textbf{52} \\ %
\bottomrule
\end{tabular}
\caption{\textbf{Spatial navigation results.} While reasoning in text may be suitable for 2D grid structures, non-grid geometries see large improvements from drawings.}
\label{tab:nav}
\end{table*}

\looseness-1 We present the results for navigation on different spatial structures from \citet{yamada_evaluating_2024} in Table \ref{tab:nav}. Consistent with \citet{yamada_evaluating_2024}, we observe that LLMs using text excel in the 2D grid setting but not other geometries, which \citet{yamada_evaluating_2024} hypothesize may be due to the grid setting 1) being more easily represented as coordinates in text than any other setting, especially as an upright `Square,' and 2) potentially having more data presented in that form on the Internet, e.g., related to tabular data, city grids, and 2D maze coding problems. We note that while humans may \textit{write} about square grids most often in text, grid cells -- which humans use to navigate physical spaces and even map conceptual spaces \cite{constantinescu_organizing_2016} -- represent space as a hexagonal grid \cite{hafting_microstructure_2005}, raising interesting questions about how spatial comprehension differs between LLMs and humans. WoT shows more consistent performance across different geometries, obviating the reliance on 2D-grid specific textual knowledge and highlighting the general applicability of the approach. This results in reduced performance on square grids, but higher performance across all other shapes, with the difficult hexagon geometry in particular seeing a jump from $8\%$ with CoT to $61\%$.

\section{In-the-Wild Examples}

\begin{figure}[h!]
    \centering
    \includegraphics[width=\linewidth]{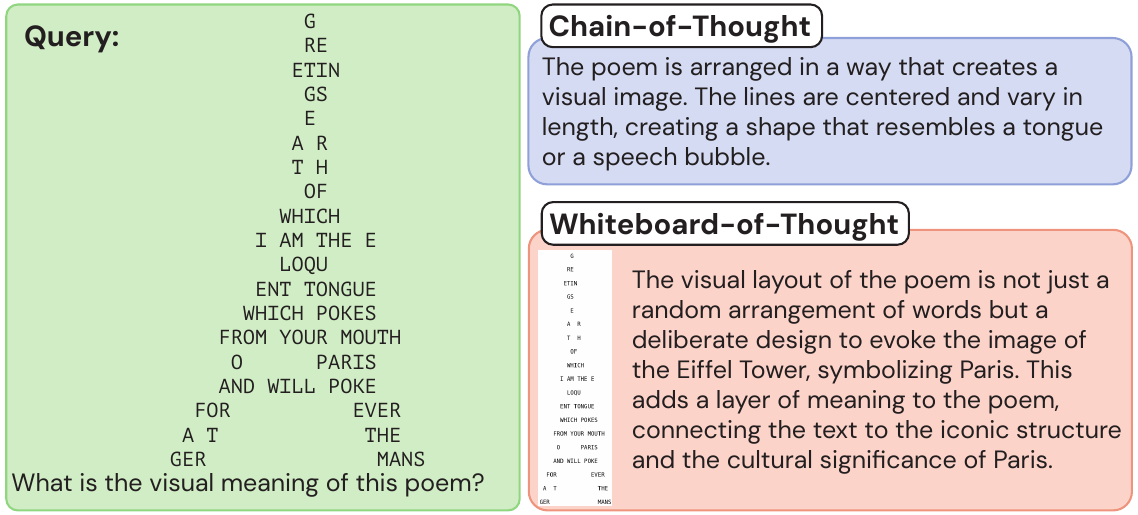}
    \caption{Calligram understanding: WoT correctly identifies the meaning behind the poem, while CoT picks up on the word `TONGUE' to hallucinate a response.} %
    \label{fig:eiffel}
\end{figure}

In this section we showcase some instances of WoT in the wild.

\paragraph{Calligrams}
Calligrams are poems where the visual organization of the words adds to the meaning of the poem. Fully understanding a calligram requires consideration of the text not only from a linguistic perspective, but also a visual one. The famous poem by Guillaume Apollinaire -- English translation: Claudia Habergham \cite{feshchenko_graphic_2019} -- in Fig. \ref{fig:eiffel} is meant to evoke the Eiffel Tower. Despite being given the same input information, CoT hallucinates the shape of tongue or speech bubble, likely due to the word `tongue' appearing in the poem. WoT, on the other hand, is able to correctly analyze the poem by first creating a visual. (Interestingly, we find GPT-4o has the original French poem memorized, and claims the words form the Eiffel Tower regardless of the actual shape given; perhaps unsurprising, as this may be one of the most famous calligrams ever made.)

\begin{figure}
    \centering
    \includegraphics[width=\linewidth]{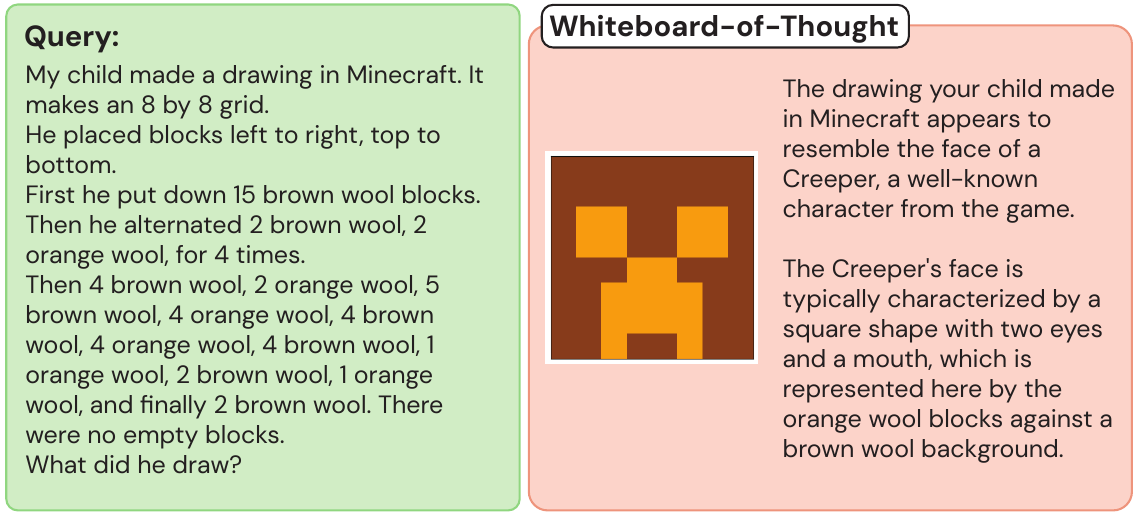}
    \caption{WoT could be used in settings like video games, where actions are expressed in text (e.g., code or logs) but have visual results.}
    \label{fig:mine}
\end{figure}

\paragraph{Video game art}
Video games and other digital spaces allow for players to interact with virtual worlds, rendered visually, through predefined actions. This has long led to players creating art through any means the game offers, often with impressive results. For games aimed at children in particular, it is an important task to detect if the visuals players create are content-appropriate. Given only text logs or description of a player's actions, it is difficult for models to process the visual implications of said actions, even though the actions fully define the resulting visuals. WoT provides a solution to this, by actually producing what the result would look like. We provide a simple example towards this goal in Fig. \ref{fig:mine}.

\section{Related Work}

\subsection{Intermediate reasoning for language models}

The success of chain-of-thought \cite{kojima_large_2023,wei_chain_2022} in arithmetic and symbolic reasoning tasks led to substantial interest in the area from the NLP community and beyond. \cite{yao_tree_2023} generalizes CoT to perform search over trees of candidate rationales. The concurrent `Visualization-of-Thought' \cite{wu_visualization--thought_2024} prompts models to produce a text pseudo-visualization, e.g., ASCII, and present results suggesting this may improve spatial reasoning as measured in the 2D grid navigation setting of \citet{yamada_evaluating_2024}; compared to our approach, this restricts the category of what can be visualized, and cannot use what the model may have learned in other modalities. \citet{zhang_multimodal_2023} and \citet{lu_learn_2022} use chain-of-thought style rationales with image and text inputs. 

All of these works, ultimately, express their intermediate results as text in some form; our work instead considers the potential of using images to express intermediate reasoning for MLLMs.

\subsection{Tool usage and code augmentation}
Scratchpads \citet{nye_show_2021} form a philosophical parallel to our whiteboards, aiming to augment a language model allowing for additional computation (in their case, a text buffer trained on Python execution traces).
PAL and PoT \cite{gao_pal_2023,chen_program_2023} achieve impressive results on arithmetic word problems by using the Python interpreter. Toolformer \cite{schick_toolformer_2023} demonstrates a training method to induce language models to invoke API calls to tools. VisProg \cite{gupta_visual_2022} and ViperGPT \cite{suris_vipergpt_2023} provide an API of visual modules to an LLM to perform visual reasoning, using a domain specific language and the Python interpreter respectively. HuggingGPT \cite{shen_hugginggpt_2023} shows LLMs can instead use existing models from the HuggingFace Transformers library \cite{wolf_huggingfaces_2020}.

\subsection{Visual and spatial reasoning in LLMs and MLLMs}

We are by no means the first to observe the limited success of LLMs and MLLMs on tasks requiring visual and spatial reasoning. The capacity of these models for grounding -- tying knowledge from the textual domain to that of other modalities, such as vision -- is controversial. \citet{patel_mapping_2022} suggests LLMs that have only seen text can perform few-shot mapping of new concepts, such as spatial directions and color, onto grounded world representations. Prompted by this, \citet{yamada_evaluating_2024} perform an in-depth evaluation of the spatial understanding of state-of-the-art LLMs, creating an extensive benchmark assessing understanding of navigation instructions on different spatial structures; they find that while GPT-4 in particular performs successful spatial navigation on 2D grids but that in general, this does not hold for other spatial structures. \citet{jiang_artprompt_2024} shows that LLMs are not capable of recognizing ASCII art which requires visual rather than simply textual understanding, and make use of this shortcoming to develop a jailbreak method that bypasses all current methods for defense and elicits unsafe behavior from state-of-the-art LLMs.

\citet{zhang_mathverse_2024,gao_g-llava_2023,kazemi_geomverse_2023} show that current MLLMs struggle to understand mathematical diagrams, such as geometric figures and graphs. Concurrent work by \citet{wang_text-based_2024} finds the same, suggesting a domain-specific language for vector graphics as an alternative input to images for LLMs to perform low-level visual reasoning. Meanwhile, \citet{huang_lvlms_2023} and \citet{han_chartllama_2023} show MLLMs perform poorly on chart understanding.
As discussed in the Limitations section, these weaknesses pose barriers to the application of WoT in these domains with current models. We are confident that as the visual capabilities of MLLMs continue to improve in these domains, WoT's performance will similarly grow.

\section{Conclusions}

We propose whiteboard-of-thought, a simple, zero-shot method to unlock visual reasoning across modalities in multimodal large language models. We accomplish this by generating code that can create a visual, then returning the visual back to the model for further reasoning. This work demonstrates whiteboard-of-thought's capabilities across multiple tasks requiring visual and spatial reasoning that have thus far proved challenging for current state-of-the-art models with text reasoning. As the abilities of these models to generate code, understand visual inputs, and perform general reasoning continue to improve, we expect the results of whiteboard-of-thought to similarly grow.

\pagebreak

\section{Limitations}

One challenge is that WoT requires accurate vision systems. As detailed in the discussion of Fig. \ref{fig:mnisterror}, a large proportion of the current errors stem from visual perception. Computer vision has advanced substantially in recent years, but there are still limitations. 

For example, a natural domain in which humans apply visual thinking is geometry; yet, even state-of-the-art MLLMs are not yet capable of understanding detailed geometric figures (see Related Work). As computer vision advances, our method will only grow more useful.

\vspace{1em}
{\small \textbf{Acknowledgments} This research is based on work partially supported by the NSF AI Institute for Artificial and Natural Intelligence Award \#2229929, the DARPA ECOLE program, the NSF CAREER Award \#2046910, and a gift from Amazon.  S.M. is supported by the CAIRFI PhD fellowship.}

\bibliography{references}

\appendix

\section{Discussion of Text-to-Image Models}
\label{sec:t2i}

Another interesting option to consider might be text-to-image models \cite{rombach_high-resolution_2022,saharia_photorealistic_2022,ho_imagen_2022}, which are already integrated into some LLMs \cite{_dall_}, or MLLMs that natively generate sequences of multiple modalities \cite{yu_scaling_2023,koh_grounding_2023,koh_generating_2023}. Text-to-image models have seen great successes, creating diverse and high-quality images from given text prompts\cite{rombach_high-resolution_2022,saharia_photorealistic_2022,ho_imagen_2022}. Despite this, it remains difficult to use them to create precise figures and drawings that would be suitable for visual reasoning, though there are exciting preliminary works in this direction \cite{rodriguez_figgen_2023,rodriguez_ocr-vqgan_2023,jain_vectorfusion_2023,xing_diffsketcher_2023}. As more effective methods for precisely controlling text-to-image models emerge, these could be easily integrated into WoT.

\section{Code and Details for Qualitative Examples}

All qualitative examples created with temperature $0$, max tokens of $2048$, top $p$ of $1$, frequency penalty of $0$, presence penalty of $0$, and with history (when the model considers the image, it also sees previous exchanges including the generated code, discussed more in Experimental Details). All generated code for qualitative figures will be released with the code for experiments. We present an example for the ASCII creation setting (from Fig.~\ref{fig:fonts}) here.

\begin{tcolorbox}[colframe=black, colback=white, title=Python Code for ASCII Art]
\begin{lstlisting}[language=Python, 
                   caption=Python Code for ASCII Art, 
                   label=code:ascii_art, 
                   basicstyle=\ttfamily\tiny,
                   keywordstyle=\color{blue},
                   commentstyle=\color{green!50!black},
                   stringstyle=\color{red}]

import matplotlib.pyplot as plt
import matplotlib.colors as mcolors
import numpy as np

# Define the ASCII art
ascii_art = [
    "".........................."",
    "".........................."",
    ""..******..*****...******.."",
    "".**////..**///**.//////**."",
    ""//*****./*******..*******."",
    ""./////**/**////..**////**."",
    "".******.//******//********"",
    ""//////...//////..////////.""
]

# Define the color mapping
color_map = {
    '.': (1, 1, 1),  # white
    '*': (0, 0, 0),  # black
    '/': (0.5, 0.5, 0.5)  # gray
}

# Determine the size of the grid
rows = len(ascii_art)
cols = max(len(row) for row in ascii_art)

# Create an array to hold the colors
image = np.ones((rows, cols, 3))

# Fill the array with colors based on the ASCII art
for i, row in enumerate(ascii_art):
    for j, char in enumerate(row):
        if char in color_map:
            image[i, j] = color_map[char]

# Create the figure and axis
fig, ax = plt.subplots(figsize=(6, 6))

# Display the image
ax.imshow(image, interpolation='nearest')

# Remove the axes
ax.axis('off')

# Show the plot
plt.show()
\end{lstlisting}
\end{tcolorbox}

\section{Experimental Details}

Here we provide details relevant to each experiment.

For all experiments, we use GPT-4o (gpt-4o-2024-05-13) as the underlying MLLM as it has each of the necessary capabilities enabling WoT -- in particular, the ability to produce code outputs and to accept image inputs. For the initial query, we use a temperature of $0$, max tokens of $2048$, top $p$ of $1$, frequency penalty of $0.05$, and presence penalty of $0$. We use these settings for all direct queries and chain-of-thought experiments as well. For the subsequent image query, we use the same settings but with max tokens set to $256$ and no frequency penalty. 

As all experiments use a temperature of $0$ and deterministic sampling, we do not perform multiple runs.

We do not use any GPU resources of our own for these experiments, instead using the OpenAI API for LLM/MLLM queries.

The core prompt enabling WoT is ``You write code to create visualizations using the \{Matplotlib/Turtle\} library in Python, which the user will run and provide as images.", provided as a system prompt. We describe some additional details here that we used to condition current MLLMs into the desired behavior. For all experiments, we append ``Do NOT produce a final answer to the query until considering the visualization." to the system prompt, as we found GPT-4o would often directly produce an answer without waiting for an image to be provided otherwise, often by hallucinating that it already had seen the image. Curiously, we found that the OpenAI content filter initially flagged a large proportion of examples, especially those that appeared pixelated and stretched to the borders, as inappropriate content. We added a white border and resized all images before sending them as queries, which somehow avoids the filter. We use the prompt from \cite{yamada_evaluating_2024}
for direct answer prompting: ``You are given a task to solve. Make sure to output an answer after "Answer:" without any explanation."

The code from the produced response for visualization typically is contained as a code block in a broader response. We extract this with the regular expression \texttt{re.findall(r"```python(.*?)```", text, re.DOTALL)}; we note this may present a source of error in failing to detect code that is not enclosed in such a block.

We found that providing both the previous history and the image in the image understanding query as opposed to passing on just the image obtained results within $1\%$ for our evaluation setting, so we elect to use the latter in general as it uses substantially fewer tokens. This choice could have interesting implications for e.g., faithfulness, where the result is guaranteed to be faithful to the image with this approach (vs not in the other case). We leave this to future work.

The ASCII understanding tasks from BIG-Bench \cite{srivastava_beyond_2022} were not included in BIG-Bench Hard \cite{suzgun_challenging_2022} due to being ``not even worth attempting with chain-of-thought." As such, we follow a similar procedure to construct evaluation splits here. In particular, we take a subset of 250 randomly chosen examples from the evaluation splits of each of the three tasks, and manually verify that they can be answered. (Running 69984 queries would also be prohibitively costly.) We remove the choices for all multiple choice questions (e.g., MNIST being given options 0-9.) For the kanji recognition task, we only include problems from the pronunciation subtask rather than the translation subtask due to lack of data. We invoke Matplotlib as the visualization tool for each of these experiments. For the image creation query, we provide the ASCII content and append each of the following: ``Write Python code with Matplotlib to render the ASCII art as an image." This avoids the model forgetting its task and directly answering. ``Let the main figure be called fig with size 6,6." This makes it easy for us to save the generated figures. ``Ensure each character in the input is considered. Remember colors are matplotlib.colors, and colors must be RGB to be displayed. Remember not all rows are necessarily the same length." These prevented common execution errors that did not seem to reflect the true visualization capabilities of the model or were due to formatting issues in the original ASCII. For the image prompt,  While we hope that as models continue to improve, these additions will be less necessary, we believe they do not detract from the results. To evaluate answers, we convert MNIST digits to integers and compare equality with ground truth, marking an instance incorrect if the type conversion fails; for word and kanji recognition, we convert the output string to lowercase and perform exact string matching. 

We use data created by \citep{yamada_evaluating_2024} to evaluate spatial understanding of LLMs for the spatial navigation experiments. In particular, we use the set described in Section 3.1 of that work, ``Do different spatial structure features affect model performance?" In particular, this includes the ring/circle, hexagon, square, rhombus, and triangle, with 100 examples for each for 500 total. We use Turtle as the visualization tool for this task. For the image creation query, we provide the navigation instructions and append each of the following: ``Use Python code with Turtle to visualize each step." As otherwise the model would still forget its instructions and produce an answer directly. ``All directions are in reference to up at setheading(90)." To provide a frame of reference for the language given. ``Name the turtle t; let the step size be 200; mark the final position with a red dot (do not write the final position as text). All other steps may be written as text." This allows us to easily save the resulting visualization. We found one limitation of the current method to be that the created visual was often correct, but had text overlap that prevented legibility (e.g., writing `Final Position' on top of the item found at that spot); this could be avoided by returning the image back to the model and asking it to modify the code, but doing this for every query is costly, resulting in these modifications. It may be interesting to explore models revising their own visuals as future work. 

\section{Potential Risks}
While this work may help against some forms of visual adversarial attacks \cite{jiang_artprompt_2024}, opening up the possibility for text queries to involve image processing could have unforeseen consequences for new forms of attacks. In addition, improving the reasoning abilities of LLMs and MLLMs in general may involve some risks, such as to certain forms of employment; however, we do not see particular ways our technique contributes to that past the general setting.

\end{document}